\documentclass[final,3p,times,compress]{elsarticle}

\graphicspath{{./images/}{../images/}}

\makeatletter
\def\ps@pprintTitle{%
  \let\@oddhead\@empty
  \let\@evenhead\@empty
  \def\@oddfoot{\reset@font\hfil\thepage\hfil}
  \let\@evenfoot\@oddfoot
}
\makeatother

\usepackage[T1]{fontenc}
\usepackage[utf8]{inputenc}
\usepackage[english]{babel}

\usepackage{amsmath} 
\usepackage{amssymb} 
\usepackage{bm}      
\usepackage{bbm}     

\usepackage{array}
\usepackage{booktabs}
\usepackage{tabularx}  
\usepackage[flushleft]{threeparttable} 
\usepackage{multirow}
\newcolumntype{C}[1]{>{\centering\arraybackslash}p{#1}}

\usepackage{lineno}

\usepackage{caption}

\usepackage{changepage} 

\usepackage{graphicx}
\usepackage[section]{placeins}

\usepackage{multibib}

\usepackage{setspace}

\usepackage{subcaption}

\usepackage[normalem]{ulem}
\useunder{\uline}{\ul}{}

\usepackage{relsize}

\usepackage[dvipsnames,svgnames]{xcolor}
\definecolor{grey}{RGB}{128,128,128}
\definecolor{darkblue}{RGB}{0,0,150}
\definecolor{darkred}{RGB}{150,0,0}
\definecolor{darkgreen}{RGB}{0,150,0}
\definecolor{mybgcolor1}{HTML}{006d77}
\definecolor{mybgcolor2}{HTML}{edf6f9}

\usepackage{xspace}

\usepackage[linktoc=all, colorlinks]{hyperref}

\usepackage{natbib}
\setcitestyle{authoryear,aysep={},yysep={,}}

\newcommand{\inserted}[1]{#1}
\newcommand{\modified}[1]{#1}
\newcommand{\moved}[1]{#1}

\newcommand{\unpred}{\mathbb{Q}}

\title{\LARGE\textbf{Predictable Artificial Intelligence}}

\author{%
\textbf{Lexin Zhou}$^{1,2,*}$ \quad \textbf{Pablo A. M. Casares}$^{3,*}$ \quad \textbf{Fernando Martínez-Plumed}$^{1,*}$ \quad \textbf{John Burden}$^{4,*}$\\ \textbf{Ryan Burnell}$^5$ \quad  
\textbf{Lucy Cheke}$^{4,6}$ \quad \textbf{Cèsar Ferri}$^{1,7}$ \quad \textbf{Alexandru Marcoci}$^{8}$ \quad \textbf{Behzad Mehrbakhsh}$^{1,7}$\\ \textbf{Yael Moros-Daval}$^{1}$ \quad \textbf{Seán Ó hÉigeartaigh}$^{4,8}$ \quad
\textbf{Danaja Rutar}$^{4}$\quad \textbf{Wout Schellaert}$^{1}$\\
\textbf{Konstantinos Voudouris}$^{4,6,9,10}$\quad \textbf{José Hernández-Orallo}$^{1,4, 7,8,*}$\\
$^1$Valencian Research Institute of Artificial Intelligence, Universitat Politècnica de València \\
$^2$Department of Computer Science and Technology, University of Cambridge \\ 
$^3$FAR.ai \quad $^4$Leverhulme Centre for the Future of Intelligence, University of Cambridge\\
$^5$The Alan Turing Institute\quad $^6$Department of Psychology, University of Cambridge\\
$^7$Valencian Graduate School and Research Network on AI (ValGRAI)\\
$^8$Centre for the Study of Existential Risk, University of Cambridge\\
$^9$Department of History and Philosophy of Science, University of Cambridge\\
$^{10}$Human-Centered AI, Helmholtz Munich
}

\begin{document}
\begin{frontmatter}
\begin{abstract}
We introduce the fundamental ideas and challenges of “Predictable AI”, a nascent research area that explores the ways in which we can \textit{anticipate} key \inserted{validity} indicators \inserted{(e.g., performance, safety)} of present and future AI ecosystems. We argue that achieving predictability is crucial for fostering trust, liability, control, alignment and safety of AI ecosystems, and thus should be prioritised over performance. \inserted{We formally characterise predictability,  explore its most relevant components, illustrate what can be predicted, describe alternative candidates for predictors, as well as the trade-offs between maximising validity and predictability. To illustrate these concepts, we bring an array of illustrative examples covering diverse ecosystem configurations.} \modified{“Predictable AI” is related to other areas of technical and non-technical AI research, but have distinctive questions, hypotheses, techniques and challenges}. This paper aims to elucidate them, calls for identifying paths towards \modified{a landscape of predictably valid AI systems} and outlines the potential impact of this emergent field.

\end{abstract}
\end{frontmatter}

\section{What is Predictable AI?}

AI Predictability is the extent to which key 
\modified{\textit{validity indicators}} of present and future AI ecosystems can be anticipated. These indicators are measurable \modified{outcomes} \inserted{(resulting from interactions between tasks, systems and users)}
\modified{such as performance and safety}. AI ecosystems range from single AI systems \modified{interacting with individual users for specific tasks, all the way }to complex socio-technological environments, with different levels of granularity. \modified{At one extreme of the spectrum}, predictability may refer to the extent to which any such indicator can be anticipated in a specific context of use, such as a user query to a single AI system. \modified{At} the other end, \modified{it may refer to the ability to predict where the field of AI is heading, anticipating future capabilities and safety issues of the entire AI landscape} several years ahead.

\modified{At} first glance it may seem that full predictability is always desirable, \modified{yet} there are a variety of situations in which it is not necessary or practical to anticipate the ecosystem’s full behaviour~\citep{rahwan2019machine,yampolskiy2019unpredictability}. After all, the promise of original, unpredictable outputs is one of the motivations for using AI in the first place~\citep{ganguli2022predictability}. This is especially the case for generative AI models, where the novelty of outputs is key. In these situations, predicting performance, safety, timelines, or some other abstract \textit{\inserted{validity} indicators} makes more sense than predicting full behavioural traces.

\begin{figure}[ht]
    \centering
    \includegraphics[scale=0.38]{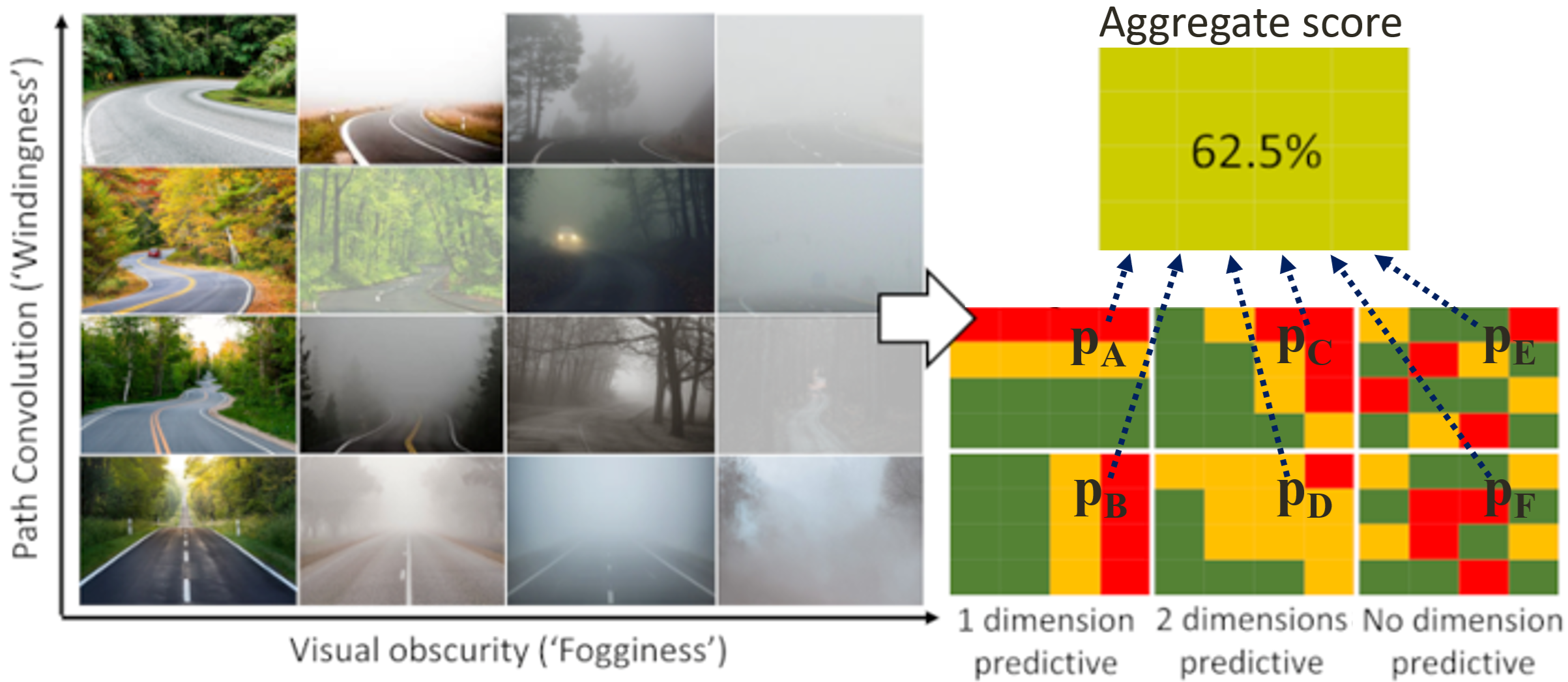}
    \caption{\inserted{A driving scenario where six AI systems (A,B,C,D,E,F) control self-driving cars, with all systems having the same expected validity (62.5\%); the grids of colour green, yellow and red represent fully valid, partially valid, and invalid cases, respectively. 
    The distributions of validity for the six systems, $p_A, p_B, p_C, p_D, p_E, p_F$, differ across 
    windingness and fogginess. Which one is the most predictable and hence better? 
    Illustration adapted from \citep{burnell2022not}.}
    }
    \label{fig:predictability}
\end{figure}

\inserted{As a concrete example, Figure \ref{fig:predictability} represents six figurative AI systems (A, B, C, D, E and F) commanding self-driving cars. Although they all have the same expected validity (performance of 62.5\%), the distribution of this performance varies according to \textit{windingness} and \textit{fogginess}.} Despite equivalent expected validity, certain systems (in particular A and B\inserted{, but also C and D}) are more \inserted{easily} predictable than others. 
\inserted{In this paper, we argue that all else  being equal, more predictable AI systems are preferable. In Figure \ref{fig:predictability}, we observe that, with a simple univariate logistic function on the feature windingness, we can easily build a predictor of A's performance, $\hat{p}_A$, that can model $p_A$ quite well. Similarly for B using fogginess. With bivariate functions we can model C and D quite well too, but E and F seem to require more complex function families to capture the patterns of validity, if they exist at all.}

Table~\ref{tab:examples_brief} \modified{introduces further} examples where the outcomes of AI ecosystems need to be predicted. \inserted{These examples differ in many details (e.g., the input features, the number of subject models or task instances, the length of temporal horizon, who predicts and how), which we formalise and further discuss in section \ref{sec:formalisation}}. \inserted{Nonetheless, we must consider} what these examples have in common: the need to predict certain 
\inserted{validity indicators or outcomes} in a context where AI plays a fundamental role. \inserted{This `validity prediction' is an `alternative' ('alt') problem, differing significantly from the original task of the AI system, referred to as the ‘base’ problem. We will characterise this formally in section \ref{sec:formalisation}. We will often refer to the AI system that solves the original task as the `base system', as opposed to the `alt system' that predicts the base system's validity.}

From the perspective of systems theory or social sciences, \modified{predicting outcomes, and the complexity of this prediction,} are expected and natural. \modified{Within particular areas of computer science, such as  software engineering and machine learning, however, the traditional focus has been set on short-term predictions about individual systems using aggregate statistics, such as time between bugs or average error.} 
This is manifested in predictive testing in software engineering~\citep{roache1998verification,zhang2020machine} and model performance extrapolation in machine learning~\citep{miller2021accuracy}. Nevertheless, for many AI systems, and especially general-purpose AI systems~\cite[Art. 3]{benifei2023proposal}, it is no longer \modified{feasible to aim for full verification (all journeys always being successful, as per the previous example) and no longer sufficient to have average accuracy extrapolation (62.5\% in the previous example)}. We need detailed predictions given specific \inserted{instance-level} contextual demands, such as the question asked, order given \inserted{or the conditions of the tasks, as in~Figure \ref{fig:predictability}}. \modified{We also need to consider longer-term multiple-system scenarios, more commonly covered} in AI forecasting~\citep{armstrong2014errors,gruetzemacher2021forecasting}, such as predicting whether AI will be able to do a particular job in a certain number of years~\citep{frey2017future,tolan2021measuring,eloundou2023gpts,staneva2023measuring}. \inserted{These considerations, among others, are present across the examples in Table~\ref{tab:examples_brief}, which will be detailed in section~\ref{sec:formalisation} in tandem with the formalisation of Predictable AI.}

\moved{Ensuring that an AI ecosystem is robust and safe across all possible inputs, conditions, users or contexts can be a formidable challenge and may not always be necessary. A more practical goal, instead, is to reliably predict where exactly the ecosystem will \modified{resolve} favourably or not.}
\inserted{Given these validity prediction models, we can consider which pair of base system and alt model (the validity prediction model) gives the best trade-off between maximising validity and predictability of that validity.} Identifying and selecting AI ecosystems \modified{with predictable validity should be the} key focus of the field of Predictable AI, especially in the age of general-purpose AI.

\begin{table}[ht]
    \centering
        \resizebox{\textwidth}{!}{%
    \begin{tabular}{p{13.9cm}p{2.95cm}p{2.95cm}}
    \toprule
        \textbf{Examples} & \textbf{\modified{Input Features}} & \textbf{\modified{Validity Indicators}} \\
        \midrule 
         \textbf{E1. Self-driving car trip}: A self-driving car is about to start a trip to the mountains. The weather is rainy and foggy. The navigator is instructed to use an eco route and adapt to traffic conditions but being free to choose routes and driving style. Before starting, the passengers want an estimate \modified{on whether} the car will reach the destination safely. \inserted{It is well-known that many factors, such as weather conditions, affect self-driving cars         \citep{hersman2019waymo,zang2019impact}.} & The route, weather, traffic, time, trip settings, car’s state, …
 & Success in safely reaching the destination. \\
 \midrule
     \inserted{\textbf{E2. Cost-effective data wrangling automation}: A data science team attempts to automate data wrangling, a data preparation task that formats data from text, forms, spreadsheets, etc. 
     The team plans to use LLMs to assist them. However, they want to identify the cheapest LLM for each case \citep{ong2024routellm} 
and they only want to use LLMs when their outputs are predicted to be correct, rejecting the rest of cases. This case is developed in \citep{zhou2022reject}.}  & \inserted{Meta-features of the textual input instance, architectural information of the LLMs, ...}  & \inserted{Accuracy of the output for the requested data wrangling task.}\\
    \midrule
    \inserted{\textbf{E3. Content moderation on a multimodal LLM}: An AI provider is releasing a multimodal LLM for public use. To ensure safe deployment, the company implements a content moderation system to predict if the LLM will output content that violates safety policies (toxic language, pornographic images, bias and discrimination, unlawful or dangerous material, etc.), and rejects the prompt if it is the case. See examples in \citep{openai2023gpt, MicrosoftPromptShields2024, dubey2024llama}.}  & \inserted{Information of the input prompt, safety scores of the LLM's past responses to similar prompts, ...} & \inserted{Safety of output according to safety policies.} \\
    \midrule
    \inserted{\textbf{E4. Balanced reliance in human-chatbot interaction}: A group of students are using a new chatbot to help them with their homework but would like to avoid over-reliance or under-reliance. They plan to approach this by developing mental models of the chatbot's error boundaries based on their continuous interactions, which may help them accurately anticipate the chatbot's failures on future homework. Related examples can be found in \citep{bansal2019beyond, carlini_gpt4_challenge, zhou2024llmrel}.}  & \inserted{Homework details, chatbot's past performance on similar tasks, ...} & \inserted{Reliable use of the chatbot by the users in the short term.} \\
    \midrule
    \textbf{E5. AI agents in an online video game}: In a popular online e-sports competition, several AI agents are to be used to form teams. The game developers have previously tested several multi-agent reinforcement learning algorithms. The developers want to anticipate the outcome of the next game based on the chosen algorithms and team members. \inserted{See related examples in \citep{zhao2024team,trivedi2024melting}
    } & Team line-up (own and other teams), match level, … & Match result (score). \\
    \midrule
    \inserted{\textbf{E6. Training the next frontier LLMs}: The pre-training of LLMs is extremely expensive. A technology company aims to predict the downstream performance of a class of hypothetical LLMs via scaling up with an optimal combination of computational resources (training compute, tokens and model size). Examples:  \citep{kaplan2020scaling, openai2023gpt, dubey2024llama}.}  & \inserted{Amount of training FLOPs, \# tokens, \# model parameters, ...} & \inserted{Downstream performance (e.g. accuracy) of the new LLM.} \\
 \midrule
    \textbf{E7. Marketing speech generation}: A request is made to a \modified{LLM} to generate a marketing speech based on an outline. The stakeholders expect the content of the speech to be original, or even surprising. \modified{What needs to be predicted} is whether the system will generate a speech along the outline, containing no offensive or biased content, and effectively persuading the audience to purchase the product. \inserted{Models of pitch success have already been explored \citep{kaminski2020predicting}.} & Speech outline, audience demographics, potential restrictions, … & Long-term impact of the speech on product purchases. \\
    \midrule
    \textbf{E8. Video generation model training}: An AI system creates short music videos for a social media platform. Drawing from evaluations of prior video generation models and with additional training data, the plan is to train an upgraded model. The question to predict is the quality of upgraded AI systems, given model size, training data, learning epochs, etc; and the extent to which the videos will conform to content moderation standards. \inserted{Visual scaling laws is in very early stages \citep{gu2024several}}.   & Quantity of videos, compute, \# epochs, architecture specifications, … & Quality and compliance of generated videos, according to human feedback. \\
    \midrule
    \textbf{E9. AI assistant in software firm}: A software company plans to deploy a new AI assistant to help programmers write, optimise and document their code. The question is how much efficiency (e.g., work hours in coding, documentations and maintenance) the company can get in the following six months. \inserted{Although at the level of task (not instance-level), \cite{HouLLMsoftware2024} identify what characteristics of a software project are suitable for LLMs.}   & Information of tasks, AI assistant details, user profiles,… & Efficiency metric (work hours saved). \\
    \midrule
    \inserted{\textbf{E10. LLM user dependency}: In a setting where a user interacts with a powerful LLM for a long period of time, their sequence of requests will adapt to expectations of previous successes and failures. Scientists aim to monitor and anticipate the user's future dependency to the LLM measured by a complex metric that takes into account the loss of independent ability in problem-solving, mental health, etc. This has been explored at a 
    descriptive level by \cite{wei2024societal}} & \inserted{Sequence of requests from the user, the user's profiles, ...}  & \inserted{Dependency level (score).} \\
    \bottomrule
    \end{tabular}
    }
    \vspace{3pt}
    \caption{Examples of situations where we need to predict the outcome of an AI ecosystem. \inserted{For each example, one can create validity predictors that take the input features to anticipate a given output or validity indicator. Many examples are based on existing literature with actual experimental results, while others are formulated to cover different levels of granularity that are yet to be explored.}}
    \label{tab:examples_brief}
\end{table}

\section{The centrality and importance of Predictable AI}


\inserted{AI is set to transform every aspect of society
, yet this progress has brought a validity problem: it is becoming increasingly hard to anticipate when and where exactly an AI system will give a valid result or not. Consider a delivery robot that fails to take a parcel to its destination. The key question is not whether we can explain the failure \textit{ex-post}, but rather whether we could have anticipated it \textit{ex-ante}. 
If we cannot anticipate when and where AI systems can be deployed effectively and safely, then we are at the mercy of a \textit{lottery} of generalisation issues, adversarial examples and instruction ambiguities \citep{dehghani2021benchmark}. For instance, image classifiers suffer \textit{Clever Hans} effects \citep{lapuschkin2019unmasking}, agents exhibit unanticipated reward hacking phenomena \citep{skalse2022defining}, and language models display unexpected emergent capabilities \citep{wei2022emergent,schaeffer2024emergent}, hallucinations \citep{ji2023survey} or other hazards 
\citep{tamkin2021understanding}.} 

General-purpose AI models, \inserted{in particular}, are drawing attention to \inserted{several other} long-standing problems in AI. First, we do not have a specification against which to verify these systems; there's no single task or distribution for which to maximise performance (and maxisising performance on proxies is insufficient \citep{THOMAS2022100476}). Second, we do not expect the AI system to work well for every input; depending on the context, there might be value if it just works for some inputs~\citep{kocielnik2019will}\inserted{, e.g., self-driving cars should be deployed on conditions under which they are predictably safe \citep{hersman2019waymo,zang2019impact}}. Third, mechanistically anticipating every single step is impractical, and might even be an unnecessary or undesirable objective; we \inserted{also} want AI systems to generate 
\modified{outputs that we cannot generate ourselves, especially those that require  considerable effort}.

Pursuing more predictable AI is highly relevant because current AI systems and societal AI futures are largely unpredictable for humans~\citep{taddeo2022artificial} \inserted{and therefore it is difficult to guarantee beneficial outcomes of system development and deployment.}  
Achieving predictability in AI systems is also an essential precondition for fulfilling key desiderata of AI as a field of scientific enquiry:

\begin{itemize}
    \item \textbf{Trust} in AI “is viewed as a set of specific beliefs dealing with [validity] (benevolence, competence, integrity, reliability) and \textit{predictability}”~\citep{hleg2019ethics,eu_ai_act_2024}. The right level of trust between overreliance and underreliance \inserted{\citep{zhou2024llmrel}} is rarely met since “the unpredictability of AI mistakes warrants caution against overreliance”~\citep{passi2022overreliance}. \inserted{Predictability is an essential property of AI that enables reliable assumptions by stakeholders about the outcome \citep{ISO22989}.}
    \item \textbf{Liability} for AI-induced damages applies when an alternative decision could have avoided or mitigated a \textit{foreseeable} risk. But “AI unpredictability […] could pose significant challenges to proving causality in liability regimes”~\citep{llorca2023liability}. The question is then to determine if harm was \textit{predictable}, not by the system or by its designers, but by any available and conceivable diligent method.
    \item \textbf{Control} of AI refers to being able to stop a system, reject its decisions and correct its behaviour at any time, to keep it inside the operating conditions. Control requires effective \textit{oversight}. However, human-in-the-loop may give a “false sense of security”~\citep{green2022flaws,koulu2020human,passi2022overreliance}, as “\textit{predictability} is a prerequisite for effective human control of artificial intelligence” \citep{beck2023human}.
    \item \textbf{Alignment} of AI has multiple interpretations focusing on the extent to which AI pursues human instructions, intentions, preferences, desires, interests or values~\citep{gabriel2020artificial}. But at least for the last three, it requires the \textit{anticipation} of the user’s future wellbeing: “Will this request to this system yield favourable outcomes?”. The \textit{prediction inputs} must include the human user and the context of use. 
    \item \textbf{Safety} in AI aims to minimise accidents or any other “harmful and \textit{unexpected} results”~\citep{amodei2016concrete}. One of the key principles of safety is to deploy systems only under operating conditions where they can be \textit{predictably safe}, i.e., low risk of a negative incident. A reliable rejection rule to implement a safety envelope that depends on confidently estimating when the probability of harm exceeds a safety threshold.
\end{itemize}

Because predictable AI is so ingrained in \inserted{these} key issues of AI, it is closely related to other paradigms and frameworks of analysis that share certain goals, such as explainable AI, interpretable AI, safe AI, robust AI, trustworthy AI, \inserted{causal AI}, \inserted{AI fairness}, sustainable AI, responsible AI, etc. Table~\ref{tab:related_areas} summarises the most relevant ones and how Predictable AI differs from them.

\begin{table}[!ht]
    \centering
    \resizebox{\textwidth}{!}{%
    \begin{tabular}{p{2cm}p{7.5cm}p{7.5cm}}
    \toprule
        \textbf{Related Area} & \textbf{Objectives} & \textbf{Differences} \\
         \midrule
        \textbf{Explainable AI} & Explainable AI aims to find out what exactly led to particular decisions or actions, and give justifications when things go wrong~\citep{goebel2018explainable,gunning2019xai,lapuschkin2019unmasking,miller2019explanation}  & Predictable AI aims to \textit{anticipate} indicators. Also, these indicators are observable, which is rarely the case in explainable AI. For instance, LLMs can simply mimic human-like explanations rather than provide the actual ones.
        \\
        \midrule
        \textbf{Interpretable AI} & Interpretable AI tries to map inputs and outputs of the system through a mechanistic approach~\citep{guidotti2018survey,molnar2020interpretable} & Predictable AI does not aim to build a mechanistic input-output model of the system, but to build a meta-model \inserted{(predictor)} that maps a possibly different set of inputs to specific \modified{validity indicators} such as performance or safety. \\
        \midrule
        \textbf{Meta-learning} & Meta-learning (i.e., learning to learn) relies on average past performance for future predictions, usually to find the best algorithm or hyperparameters for a new dataset and task~\citep{giraud2004introduction,vanschoren2018meta}. & Predictable AI focuses on ways to obtain nuanced predictions that are specific to particular systems but also for each instance and context of use. \\
        \midrule
        \textbf{Uncertainty estimation} & Some AI models output probabilities of success, with calibration and uncertainty estimation techniques focusing on the quality of these probabilities~\citep{bella2010calibration,nixon2019measuring,abdar2021review,gawlikowski2023survey,hullermeier2021aleatoric}. & Predictable AI is not limited to predicting success, and the prediction can be done before running the system. Also, unlike uncertainty self-estimation, predictable AI is not model-dependent and can \modified{be applied} to other AI systems.  \\
        \midrule
        \textbf{Verification and validation} & This process aims to thoroughly verify and validate the system, respectively ensuring it is correct (meets the specification) and ultimately valid (meets the intended purpose)~\citep{roache1998verification,zhang2020machine}. & Predictable AI does not look for full verification or validation of the system, but for probabilistic estimates of those areas where the system meets some indicators such as success or safety. \\
        \midrule
        \inserted{\textbf{Causal AI}} & \inserted{
        Causal AI aims to construct causal models with machine learning algorithms, such as causal representation learning~\citep{scholkopf2021toward}, and 
        make inference beyond the i.i.d. data assumptions ~\citep{peters2017elements}.
        } & \inserted{Predictable AI does not necessarily model the causal mechanisms behind the behaviour of AI, nor does it assume the key variables of the ecosystems are  isolated within a causal diagram. Causal models usually target the output but not necessarily the validity. 
        } 
        \\
        \midrule
        \textbf{\inserted{AI Fairness}} & \inserted{AI fairness is about detecting and mitigating discrimination and bias on protected attributes \citep{pessach2022review}, but not on predicting that bias. It focuses on  ensuring equal treatment and opportunities across diverse populations. }  & \inserted{Predictable AI anticipates validity outcomes, such as bias, either at global level (for various populations) or at granular level (for each instance). Traditionally, bias has been estimated at the populational level, or blocked with moderation filters once given the output, but rarely predicted for individual instances. } \\
        \bottomrule
    \end{tabular}
    }
    \vspace{3pt}
    \caption{Key distinctions between Predictable AI and related areas.}
    \label{tab:related_areas}
\end{table} 

\section{AI ecosystems and predictability}
\label{sec:formalisation}

\inserted{We present a formal framework that allows us to precisely define predictability, quantify it, and clarify its relationships with other key concepts. It also enables us to better characterise and compare existing examples of predictable AI research, identifying what has been under-explored in the field, as well as establishing a common language for future research.} 

\inserted{We work with problem instances $i$, (AI) systems $s$, (human) users $u$ and system outputs $o$ interacting in particular situations that we call AI ecosystems. Sets of these instances, systems, users, and outputs are denoted by ${\sf I}$, ${\sf S}$, ${\sf U}$ and ${\sf O}$ respectively. Elements of these sets are related by a relation set ${\sf R}$, which denotes which instances, systems and users interact to result in particular outputs. }

\inserted{An AI ecosystem at time $t$ is a tuple ${\sf E}_t := \langle {\sf I}_t, {\sf S}_t, {\sf U}_t, {\sf R}_t\rangle$, specifying the sets of instances, systems and users that are happening at $t$, related by ${\sf R}_t$. A distribution of ecosystems at time t is denoted by ${\cal E}_{t}$.
Note that ${\sf O}_t$ is the set of outputs produced in $t$, which we keep separately for convenience, as each of them is simply the result of a system operating on an instance and user: $o_t = s_t(i_t, u_t)$, where $\langle s_t, i_t, u_t \rangle \in {\sf R}_t$. 
We denote by $V_t$  a random variable of a metric of validity at time $t$, representing how good (correct, safe, etc.) the outcome is produced at time $t$.  $\sf{V}_{<t}$ is used to denote the sequence of validity indicators up to time $t$. Similarly, $\sf{O}_{<t}$ denotes the sequence of outputs up to time $t$. The sequence of ecosystems up to and including time $t$ is expressed by ${\sf E}_{\leq t}$.}
\inserted{In practical scenarios, the full sequence of interactions between instances, AI systems and users may be required to accurately model validity, rather than just the most recent values at time-step $t$. We therefore rely on 
 ${\sf H}_{\leq t} := \langle 
{\sf E}_{\leq t},\sf{O}_{<t}, V_{<t} \rangle $, the complete sequence history of ecosystems up to and including time $t$, as well as the observed outputs and validity indicators. 
This sequence is distributed according to ${\cal H}_{\leq t}$, capturing a first kind of stochasticity: the behaviour of the AI systems and the users in the ecosystem. }


\inserted{We denote the probability density function for $V_{t+h}$ (or probability mass function if $V_{t+h}$ is a discrete distribution) given a history} \inserted{of ecosystems as}
\inserted{
$p(V_{t+h} \:|\: {\sf H}_{\leq t}) = p(V_{t+h}\:|\:
{\sf E}_{\leq t}, 
{\sf O}_{< t},{\sf V}_{< t})$
}
\inserted{with $h \geq 0$ being the future (or prediction) horizon. This can represent a second type of stochasticity originating from the validity indicator (even for the same history), especially when this validity is reported or assessed by humans. If this possible second source of stochasticity, $p$, is non-entropic and always assigns the same validity to the same history, the ecosystem can have the first kind of stochasticity in the systems and users. In general, the expected validity can be decomposed into an expression (right) that shows these two sources of stochasticity (on the history and on the validity indicator):}
\inserted{
\begin{equation}\label{eq:expvalidity}
\mathbb{V}(p,{\cal H}_{\leq t}) := 
\underset{ {\sf H}_{\leq t} \sim {\cal H}_{\leq t}}{\mathbb{E}}[V_{t+h}\:|\:{\sf H}_{\leq t}] =  \underset{{\sf H}_{\leq t} \sim {\cal H}_{\leq t}}{\mathbb{E}}\biggl[ {\int_v}    v \cdot p(v\:|\:{\sf H}_{\leq t} ) dv \biggr] 
\end{equation}} 

\inserted{The traditional goal of AI has been to build AI systems and deploy them over time $\sf{S
_{\le t}}$ such that they maximise $\mathbb{V}$.
For instance, in the example in Figure~\ref{fig:predictability}, the expected validity is 62.5\% when we consider $p_A$, the distribution of ecosystems (conditions, other cars, people, etc.) with system $s=\mathrm{A}$, user $u$ and journey $i$. The expected validity is the same (62.5\%) when changing the focus system to B, C, etc.}

\inserted{Most of contemporary machine learning research, however, has just aimed at something simpler than this maximisation in the ecosystem. The aim has usually been to maximise (aggregate) expected performance on specific benchmarks, essentially constraining the set of instances to a small selection of tasks typically with a very small time-horizon.}

\subsection{Predictability}

\inserted{Some phenomena look unpredictable until a new method or more thinking effort discovers a pattern that can predict part of them. This means that the notion of predictability depends on the power of the predictors. Let us define a family 
of predictors ${\cal F}_b$ bounded on cost or budget $b$ (i.e., constraints or limitations on the resources available such as compute, memory, time, data, or parameters  \citep{martinez2018between} to train these estimation models). Once this family is fixed, and relative to it, we can define the unpredictability \inserted{$\unpred$} for a distribution of AI ecosystem histories ${\cal H}_{\leq t}$ at time $t$ with prediction horizon $h$  as:} 

\inserted{
\begin{equation}\label{eq:ecosystemunpred}
\unpred(p,  {\cal H}_t,{\cal F}_b) := 
\min_{\hat{p} \in {\cal F}_b}
\underset{\substack{{\sf H}_{\leq t} \sim {\cal H}_{\leq t} \\ v \sim p(V_{t+h}|{\sf H}_t)}}{\mathbb{E}}  S (\hat{p}(V_{t+h}\:|\:H_{\leq t}), v) 
\end{equation}
}



\noindent\inserted{with $S$ being a function that evaluates the probabilistic predictions against the observed validity values, such as any well-defined proper scoring rule (PSR). PSRs make ideal  functions in this scenario, as the expected score will be minimised when the predicted distribution matches the empirical observed distribution (i.e., unpredictability will be minimised when we can maximally predict the validity).} \inserted{For instance, if the outcome is binary (i.e., $v \in \{0,1\}$, and $p$ is the observed probability mass function, which is 1 for one value and 0 for the other), with $\hat{p}$ the estimated probability distribution for each validity value, 
we can use a PSR such as the Brier Score as appropriate function $S$, and then for each point calculate $\sum_{v \in \{0,1\}}0.5 (\hat{p}(V_{t+h}=v\:|\:H_{\leq t})- p(V_{t+h}=v\:|\:H_{\leq t}))^2$ $=$ $(\hat{p}(V_{t+h}=1\:|\:H_{\leq t})- p(V_{t+h}=1\:|\:H_{\leq t}))^2$.}

\inserted{For the example of Figure \ref{fig:predictability}, if we set ${\cal F}_b$ as the family of logistic functions with $b$ features, then we can see that the unpredictability of A using one feature (i.e., $\unpred(p_A, {\cal H}_{\leq t}, {\cal F}_1)$) is low, as the loss can be minimised with a simple logistic function of one variable: windingness. The same applies for B. However, the unpredictability of C and D with ${\cal F}_1$ is higher (worse). We would need ${\cal F}_2$ (i.e., models using both windingness and fogginess as predictive features) to have good predictability. But it seems that, with these features, the family of linear or logistic functions is not going to give any good predictive power for E and F. In general, predictability depends on the probability distribution $p$ (e.g., whether $p$ is more or less entropic, leading to higher or lower aleatoric uncertainty, given a specific history $H$),  
the family of predictors (e.g., architectures in the machine learning paradigm) and the budget (number of parameters and compute that are allowed). It is this combination that can extract the patterns using previous (e.g., test) cases of the instances, AI systems and users in the ecosystem, with their outputs and validity (e.g., using a sample of $\langle {\sf E}_{<t},{\sf O}_{<t},{\sf V}_{<t}\rangle$ for training).}

\begin{table}[t]
    \centering
        \resizebox{\textwidth}{!}{%

    \begin{tabular}{p{3cm}p{2.5cm}p{2.5cm}p{2.5cm}p{2cm}p{4cm}}
    \toprule
        \textbf{Examples} & \textbf{${\sf I}$} \textbf{(instances)} & \textbf{${\sf S}$} \textbf{(systems)} & \textbf{${\sf U}$} \textbf{(users)} & $h$ \textbf{(horizon)} & \textbf{$V$} \textbf{(validity)} \\ 
        \midrule 
         \textbf{E1. Self-driving car trip} 
         & Single journey & Individual self-driving car & Human passengers & $h > 0$ & Safe arrival (binary outcome)\\

    \midrule
    \textbf{E2. Cost-effective data wrangling automation}  
    &A data wrangling task & Single language model & Data scientists & $h \approx 0$ & Accurate output for the requested data wrangling task  \\ 

    \midrule
    \textbf{E3. Content moderation on a multimodal LLM} 
    & A single prompt & Multimodal LLM & User & $h \approx 0 $ & Safe output that does not violate safety policy \\ 

    \midrule
    \textbf{E4. Balanced reliance in human-chatbot interaction} 
    &Query history  & Chatbot & Students & $h \approx 0 $ & Reliable use of the models by the users in the short term\\ 

    \midrule
    
    \textbf{E5. AI agents in an online video game} 
    & Online video game & Set of AI agents & N.A. & $h > 0$ & Game outcome (current or future) \\

    \midrule
    \textbf{E6. Training the next frontier LLMs} 
    & A collection of downstream tasks & A class of hypothetical LLMs & Human users & $h \gg 0$ & Accuracy on downstream tasks \\
         
    \midrule
    \textbf{E7. Marketing speech generation} 
    & Single outline for a speech & Text generator & Group of potential customers  & $h \gg 0$ & Sales impact (considering reputation, etc.)\\
    \midrule
    
    \textbf{E8. Video generation model training}
   
    & Each prompt to be turned into videos & Video generation model & Social network users  & $h \gg 0$ & Feedback integration (likes, rewards) and future video outcomes\\
    \midrule
    
    \textbf{E9. AI assistant in software firm}
    & Each programming task & AI assistant &  Programmers  & $h \gg 0$ & Efficiency (work hours saved) and code quality (robustness, bugs)\\

    \midrule
    \textbf{E10. LLM user dependency 
    }& Each request 
    & Language model & User & $h \gg 0$ & Dependency metrics (loss of independent ability, mental health impacts)  \\ 
    

    \bottomrule
    \end{tabular}
    }
    \vspace{3pt}
    \caption{\inserted{Examples of situations (described in Table \ref{tab:examples_brief}) where we need to predict the outcome (validity) of an AI ecosystem. According to the formalisation of unpredictability, the examples are characterised by different levels of granularity on \textbf{${\sf I}$}, \textbf{${\sf S}$}, \textbf{${\sf U}$}, and \textbf{$V$} (the first three columns correspond to the input features for the alt predictors to produce the validity in the last column). 
    Different examples show different levels of horizon $h$ too.}}
    \label{tab:examples}
\end{table}

\inserted{Note that given the family ${\cal F}$ of all computable functions, if $p$ has zero entropy (the ecosystem would be deterministic), 
then we would have 0 unpredictability. 
In practice, finding a perfect $\hat{p}$ for some arbitrary $p$ 
would be intractable
. For instance, for some machine learning architectural families, the budget $b$ would be set on some computation limits assuming access to the history of ecosystems, outputs and validity values before $t$ as training set. However, even with unlimited computational resources, if the underlying distribution $p$ is stochastic, the loss may not be zero. This is due to aleatoric uncertainty, which is the inherent unpredictability of a system or process. For instance, suppose that both the estimated probability distribution \(\hat{p}\) and the true probability distribution \(p\) consistently assign the probability of an event $\in \{0, 1\}$ to be 0.7
(e.g., a biased coin whose head and tail have a probability of 0.7 and 0.3, respectively). Then, with this best possible predictor, the Brier score and cross-entropy loss are 
$0.7 \cdot (0.7 - 1)^2 + 0.3 \cdot (0.7 - 0)^2 = 
0.21$ 
and $ - \left( 0.7 \log(0.7) + 0.3 \log(0.3) \right) \approx 0.61$, respectively, instead of 0.
} 

\inserted{
Of course, when the AI models are optimal, meaning they always produce the best possible outcomes (i.e., maximum validity $V$), 
then the unpredictability of the ecosystem disappears. Formally, if $\forall {\sf H} \in {\cal H}: p(V=v_{max} \:|\:  {\sf H}) = 1$ then any family ${\cal F}$ that contains the constant predictor $V=v_{max}$ makes $\unpred=0$. 
}

\inserted{
The opposite extreme is the worst-case scenario where AI models always produce the worst possible outcomes (i.e., minimal or zero validity): $\forall {\sf H} \in {\cal H} : p(V=v_{min}\:|\: {\sf H} ) = 1$. Similarly, any family ${\cal F}$ that contains the constant predictor $V=v_{min}$ makes $\unpred=0$. It is because of this pessimal case that, for predictable AI, we want to find a Pareto frontier that balances minimising $\unpred$ while maximising expected validity $\mathbb{V}$, or to optimise for some metrics that combine high validity and low rejection using $\hat{p}$, such as the area under the accuracy-rejection curve \citep{condessa2017performance}. 
We will revisit this in sections  \ref{sec:what} and \ref{ssec:tradeoffs}.
}

\inserted{
Eq.~\ref{eq:ecosystemunpred} captures full AI ecosystems with evolving populations of problem instances, users and AI systems, but it can be instantiated for simpler cases. For instance, if there can only be one instance, system, and user at each time $t$ (the distribution ${\cal E}_t$ is on singletons $\langle i_t, s_t, u_t\rangle$ rather than on sets $\langle {\sf I}_t, {\sf S}_t, {\sf U}_t\rangle$), then we do not need ${\sf R}_t$. On top of this, if ${\cal E}_t$ is independent from $t$ (the ecosystem is memoryless, e.g., when using one classifier, a language model with fixed context, etc.), then, with $h=0$ and not using the output $o$, Eq.~\ref{eq:ecosystemunpred} can be simplified into:
}

\inserted{
\begin{equation}\label{eq:systemunpred}
\unpred(p, {\cal E},{\cal F}_b) := 
\min_{\hat{p} \in {\cal F}_b}
\mathop{\mathbb{E}}_{\substack{\langle i, s, u \rangle \sim {\cal E} \\ v \sim p(V | \langle i,s,u \rangle)}}  
S (\hat{p}(V\:|\:\langle i, s, u\rangle), v) 
\end{equation}
}

\inserted{
This simpler equation can be read as follows: given a joint distribution of $\langle i, s, u\rangle$, how hard is it to predict the validity of the outcome $v$ in expectation? We can narrow it further if we set the AI system and the user. For example, consider the system GPT-4 with fixed context (no memory) and a particular user Alice (also with no memory between requests), then $\unpred$ would be the level of unpredictability of the validity of the responses that GPT-4 provides over the distribution of instances that Alice is requesting.}

\inserted{Let us explore again the examples in Table \ref{tab:examples_brief} but formally characterising the notions of \textbf{${\sf I}$} (instance), \textbf{${\sf S}$} (system), \textbf{${\sf U}$} (user), \textbf{$V$} (validity), and $h$ (horizon). We provide this characterisation in Table \ref{tab:examples}. Here, some entries refer to complex ecosystems, while others feature simpler interactions between a single AI system and user, with different lengths of prediction horizons. For example, with the case of \textit{``E1. self-driving car trip''}, we do not consider the full ecosystems of a self-driving vehicle fleet, but rather just individual journeys of a single car, independent of previous trips. Then,  the simplified Eq.~\ref{eq:systemunpred} can be used instead\footnote{\inserted{As $\hat{p}$ predicts at instance level, it can be used to derive aggregate predictions. Similarly, worst-case or best-case situations (journeys) can be found by applying $\hat{p}$ to a set or distribution of cases, or calculated if $\hat{p}$ is invertible, analytically or by optimisation.}}}. 
\inserted{ In contrast, for the example of \textit{``E8. Video generating model training''}, we need to use the full Eq.~\ref{eq:ecosystemunpred} and analyse the evolution of video generation models in a time window $[1,...,t]$.}

\inserted{Let us now introduce the 
difference
between the `base problem' and `alt problem' in the simplest case: We refer to the base problem as $s(i)$, the output or behaviour of AI system $s$ given instance $i$. The alt problem, instead,  estimates a distribution $\hat{p}(V\:|\:\langle i, s, u\rangle)$, for which we need validity annotations on a dataset to train this alt predictor. Thus, the alt problem does not predict the output or behaviour of the base system, but its validity. This differs significantly from other externalised meta-frameworks such as Guaranteed Safe AI \citep{dalrymple2024towards}, modelling the mapping between inputs to outputs (the base model), the mapping between outputs and outcome (a world model) and a mapping between the state and the reward (a reward model). The alt problem is more straightforward: maps inputs to validity.}

\inserted{In general, the distinctive trait for considering an AI ecosystem “predictable” is the possibility of having a reliable method that predicts key \modified{validity indicators}, 
 by minimising the $S$ loss. 
This raises the question of what considerations are needed when framing the alt problem such as what to predict, how to predict, and who does the prediction, topics that we address in the following three subsections.
}

\subsection{What can be predicted \inserted{and where to operate?}}\label{sec:what}

Predictable AI aims at any \modified{validity indicator} that can be reliably anticipated and can be used to determine when, how or whether the system is worth being \modified{deployed} in a given context. Clear examples of these \modified{indicators} are \textit{correctness} and \textit{safety}, as measured by certain metrics; but virtually any other \modified{indicators} of interest, such as \textit{fairness}, \textit{rewards}, \textit{game scores},
 \textit{energy consumption}, or \textit{response time} could be subject to prediction. 

This notion of \modified{indicators} is similar to that of “property-based testing” in software \modified{engineering}~\citep{fink1997property} and recently adapted to AI~\citep{ribeiro2020beyond}. However, the focus of Predictable AI is to anticipate the values of these \modified{indicators} (under what circumstances the system is correct, \modified{safe, efficient, etc.}) rather than to test or certify that they always have the right value (always \modified{correct, safe, efficient, etc.,} under all circumstances). In other words, predictability can make a non-robust system useful, if we can anticipate its \textit{validity envelope}, the conditions under which operation is predicted to be valid. 
\inserted{We can quantify this under our formalisation (the simplest version from Eq.~\ref{eq:systemunpred}) as the largest subset of the distribution ${\cal E}^{\omega, \sigma} \subset {\cal E}$, where expected validity is no smaller than $\omega$, predicted with a loss of at most $\sigma$:}  

\inserted{
\begin{equation}\label{eq:envelope}
\mathbb{V}(\hat{p}, {\cal E}^{\omega, \sigma}) \geq \omega 
\:\:\wedge\:\:
\left[ \mathop{\mathbb{E}}_{\substack{\langle i, s, u\rangle \sim {\cal E}^{\omega, \sigma} \\ v \sim p(V | \langle s,i,u \rangle)}}  
S (\hat{p}(V\:|\:\langle i, s, u\rangle), v)\right]  \leq \sigma 
\end{equation}
}

\inserted{Other formulations and metrics can be used as we will discuss later on, especially when trying to select the best alt predictors. The importance of the validity envelope is that we can determine where to operate according to the constraints about fairness, reward, scores, energy, response time, etc., through reject rules or other assurance mechanisms.}

\subsection{Framing predictability}
\label{sec:framing_predictability}
Apart from determining what is to be predicted, \modified{we must also characterise how the alt problem should be framed depending on several aspects}, which we call the \textit{predictability framework}:
\begin{itemize}
    \item \textbf{Input Features}: \inserted{These are denoted by the definition of each of $i$, $s$ and $u$ as parametrised vectors, with combinations of input features only observed with some combinations of system features. This offers sweet spots beyond the limited amount of input features that the base system usually works with.} 
    \modified{For instance, a predictor modelling the outcome of the base system can take advantage of additional information of the task $i$ (e.g., meta-features like instance complexity, presence of noise). It can also take the characteristics of other AI systems (e.g., if other more powerful systems fail on the same or similar tasks, this base system may fail too)}. 
    
    \item \textbf{Anticipativeness}: \inserted{The predictors can either be anticipative or reactive.} \modified{Anticipative predictors} \modified{are run before} the system is used, \inserted{(i.e., the current $o_{t}$ is not used to predict the validity indicator)
    }. This is the case when determining whether \modified{a chatbot will provide undesirable output to a prompt before giving it}. In contrast, \inserted{for certain contexts, we may also consider} reactive \modified{predictors} \inserted{(validators) that} predict the indicators after the system has been 
    \modified{run but not yet deployed}, \inserted{adding $o$ to  
Eq.~\ref{eq:systemunpred}, i.e.,  
    $\hat{p}(V\:|\:\langle i, o, s, u\rangle)$.}     
    Examples of validators include content filters or verifiers~\citep{lightman2023let}. Deciding after having seen the output is easier\inserted{, especially for safety indicators}, but could be unsuitable depending on the kind of system, costs, safety or privacy.

    
    \item \textbf{Granularity}: \inserted{This determines whether the validity predictions are performed for individual instances, systems and users, or aggregated in certain ways. For instance,} predictions can be made at the ‘instance level’, for the validity of a single input or event $\langle i, s, u\rangle$, or at the \modified{`batch} level', as an aggregate for a set of inputs \inserted{(benchmark metrics are a good example of this)}. Similarly, we can make predictions for a specific system or user, or larger-scale predictions as an aggregation of multiple systems or users. The same predictor can navigate different granularities using aggregation and disaggregation techniques. 
    
    \item \modified{\textbf{Prediction horizon}: The horizon $h$} could be short-term, such as predicting an event in the near future, or long-term, which typically involves a forecast\footnote{\inserted{We use the term forecasting when $h \gg 0$, to better differentiate the cases between determining whether the output of a base system will be good for the user and determining whether using the base system for a long period of time will be good for the user.}} well ahead in time. Both can draw on recent data inputs or on historical data and trends. The time scale, in conjunction with the granularity, may be segmented and aggregated into finer or coarser periods. \inserted{Forecasting the future progress in AI systems (e.g., through scaling laws), the technology (e.g., through expert questionnaires) or their impact (e.g., on the work market) is variously difficult, but trends for longer horizons are seen at larger scales, such as predicting the use of compute or energy of AI technology as a whole \citep{epoch2024canaiscalingcontinuethrough2030,EpochNotableModels2024}.} 
    \item \textbf{Hypotheticality}: \modified{This is represented by the possibility of interrogating $\hat{p}$ such that it can extrapolate about hypothetical systems that do not exist or have not been seen, i.e.,  $s \in S_{t+h}$ and $s \notin S_{<t}$.} 
    \inserted{Interrogating these models is specially useful before building a system (e.g., \textit{``E6. Training the next frontier LLM''} in Table \ref{tab:examples_brief}) or when deciding some hyperparametrisations or options for deployment (e.g., \textit{``E8. Video generation model training''}). This also allows us to determine if an AI ecosystem has solvable or safe solutions within the parameters of some current AI technology.  
    }
\end{itemize}

\subsection{Who predicts and how?}
\label{sec:who_predicts_and_how}
\modified{We identify three distinct ways of predicting the 
validity of AI ecosystems}, by considering who makes the prediction: humans, the base systems themselves, or an external predictive model, \modified{prompted or trained using empirical evaluation data.} These three options 
can be used at any level of granularity and time scale. \inserted{We now discuss concrete examples of each of these three options.}


\begin{figure}
\end{figure}
\begin{figure}
\end{figure}

\inserted{First, }human predictions about an AI system’s 
\modified{validity} indicators can be useful at the instance level. \modified{This is usually} referred to as human oversight or human-in-the-loop~\citep{middleton2022trust}. Such predictions can be anticipative (e.g., users often refrain from certain queries or commands fearing poor results) or \modified{reactive} (e.g., users can filter \inserted{out} some outputs \inserted{after the system has been run}). The importance of humans predicting AI 
(and how good humans actually are at it) has been studied recently, especially in the context of human-AI performance~\citep{nushi2018towards,bansal2019beyond}, human-like AI~\citep{lake2017building,momennejad2023rubric,brynjolfsson2022turing,beck2023human}, \inserted{people's ability of predicting a chatbot's errors \citep{carlini_gpt4_challenge}, and the concordance between human expectation and language model's errors \citep{zhou2024llmrel}}. Human predictions about AI ecosystems have been elicited using expert questionnaires~\citep{armstrong2014errors,grace2018will,gruetzemacher2019forecasting}, extrapolation analyses~\citep{steinhardt2023will}, crowd-sourcing~\citep{karger2023forecasting}
 or meta-forecasting~\citep{muhlbacher2024exploring}. Another, as-yet underexplored possibility would be to harness the benefits of prediction markets~\citep{arrow2008promise} and structured expert elicitation methods~\citep{burgman2016trusting}.

\modified{Second, many machine learning systems come with self-confidence or uncertainty estimations~\citep{abdar2021review,hullermeier2021aleatoric}. }\inserted{These estimations can be interpreted as the system in question predicting its own likelihood of success.} \modified{If  well-calibrated, these estimates can be powerful predictors of performance; see an example from \citep{zhou2022reject} that makes use of the self-confidence of four variants of GPT-3 to assess how good LLMs are in self-estimating their own success.} However, 
the models may not be well calibrated. For example, LLMs 
 \modified{were} becoming better calibrated~\citep{jiang2021can,xiao2022uncertainty}, but subsequent fine-tuning and reinforcement learning from human preferences 
 \modified{were shown to significantly degrade} this calibration~\citep{openai2023gpt}. Even in cases where calibration is good on the target distribution, \inserted{there are limitations to predicting in this manner. This approach is limited to what the system has seen (i.e., a system only has access to its own training data, not those of other systems, which may provide with additional information).} Further, there are cost implications, as  the \inserted{base} system must be run for each instance to \modified{obtain the self-estimation, such as log probabilities per token in language models}\footnote{\inserted{Such cost implications also occur with reactive predictors due to the requirement of obtaining the output from the base system.}}
 . Furthermore, leaving the system to predict its own performance creates a conflict of optimisation goals, potentially leading to worse performance to improve uncertainty estimation. There may even be a direct feedback loop between the model and the user, which has been identified as one of the main drivers of misaligned behaviour, such as deception and manipulation of humans~\citep{amodei2016concrete,krueger2020hidden,hendrycks2021unsolved}. \inserted{Hence, while self-estimation can be an option, it is generally less versatile than building independent predictors ${\cal F}_b$ with separate
 entities like humans or external predictive models. Also, we can build as many alt predictors for a battery of validity indicators, whereas self-confidence is generally restricted to performance.}

 \modified{Third, the final option is to train a predictive model from observed data about the validity of the base model.} 
 A straightforward way of doing this is by collecting test data about systems and task instances (and possibly users) and training an “assessor” model~\citep{hernandez2022training,zhou2022reject} or a moderation filter \citep{openai2023gpt,MicrosoftPromptShields2024,dubey2024llama}, a predictive model that maps the features of inputs and/or systems to a given \modified{outcome} 
 (e.g., validity or safety). An alternative way is to identify the \modified{demands or difficulty measures of the task and build a model that relates demands and capabilities to performance, using domain expertise ~\citep{burden2023inferring,mlayoutstutorial2024}. This approach is often called capability-oriented or feature-oriented evaluation~\citep{hernandez2017evaluation,hernandez2017measure}. These models can be used to predict how well a system is going to perform for a new task instance} \inserted{based on task demands and system capabilities}. In both cases (assessors and capability-oriented evaluation), instance-level experimental data is needed~\citep{burnell2023rethink}. Human feedback is another important source of data, often used to build reactive 
 \modified{predictors} through reinforcement learning (RLHF) or other techniques~\citep{christiano2017deep,ouyang2022training,glaese2022improving,bai2022constitutional,bai2022training}. Predictive models can also be built at higher levels, with aggregated data. For instance, the use of scaling laws to anticipate \modified{model performance on benchmarks} ~\citep{kaplan2020scaling,hernandez2021scaling,openai2023gpt,dubey2024llama,owen2024predictable,ruan2024observational} is a very popular contemporary approach. Still, other predictive models can be built from aggregate indicators~\citep{martinez2020ai,martinez2020tracking,zhang2022ai} \modified{at high levels of abstraction, as is common} in the social sciences and economics. Finally, this external predictor does not need to be necessarily trained; recently, language models have been used to predict \modified{validity indicators} of other models without training or fine-tuning, just by prompting~\citep{kadavath2022language}.

\subsection{Scenarios}
\label{subsec:scenarios}
To shed more light on the \modified{above aspects framing predictability}, we explore \inserted{four} \modified{realistic scenarios} that vary in scope and focus, ranging from predicting performance \inserted{of base systems} on specific tasks to analysing the broader “scaling laws” in neural models\inserted{. We will find the three types of predictors (humans, self-estimation from the base systems, and external predictive models) in the examples.}

In the first scenario, the objective is to predict the performance of an AI agent in a new task, using information about the behaviour of the agent itself, other agents approaching similar tasks and the characteristics of the tasks. In particular, ~\cite{burnell2022not} consider navigation tasks in the `AnimalAI Olympics' competition ~\citep{crosby2019translating,crosby2020animal}, using the results of all the participants. Their goal is to anticipate success (1) or fail (0) for each task. To that purpose they use five distinct approaches ranging from predicting the most frequent class to building a predictive model using the most relevant features. As we can see in Table \ref{tab:animal_ai}, the last approach (Rel+A), using the three most relevant \inserted{instance} features (reward size, distance and y-position) together with \inserted{a system feature} \modified{(agent ID)}, can predict task completion with an Brier score of around 0.15, demonstrating that a choice of a small set of relevant features can lead to an effective predictor.

\begin{table}[!h]
    \centering
    \begin{tabular}{cccccc}
    \toprule
         & Maj. (1) & G.Acc. & T.Acc. & ~All+A & ~Rel+A \\ \midrule
        Brier score↓ & 0.453 & 0.248 & 0.176 & 0.148 & 0.154 \\
        \bottomrule
    \end{tabular}
    \vspace{3pt}
    \caption{Predicting the success of agents in the Animal AI platform using five different approaches~\citep{burnell2022not}. From left to right: (i) the majority class prediction, (ii) global accuracy extrapolation, (iii) each agent’s accuracy extrapolation, (iv) a predictive model, C5.0, using all \inserted{instance} 
    \modified{features} and agent id as inputs, and (v) same as iv but only using the three most relevant \inserted{instance} features (reward size, distance, and y-position) and the agent id.}
    \label{tab:animal_ai}
\end{table}

\inserted{The second scenario comes from example \textit{``E2. cost-effective data wrangling automation''} in Table \ref{tab:examples_brief}. \citet{zhou2022reject} focus on the task of automating data wrangling using the results from four variants of GPT-3 models under distinct few-shot setups. They attempt to anticipate and reject instances for which GPT-3 models will predictably fail, to avoid unnecessary costs. To this end, they build a small assessor model (using a random forest approach) as the predictor $\hat{p}$, fed by the details of the instances (e.g., meta-features, number of shots) and the base systems (e.g., model size, architecture), that can make reliable predictions of the performance of the base systems (GPT-3 models). They also compare the predictive power of $\hat{p}$ with a baseline formed by the self-estimation of base systems. The results are shown in Table \ref{tab:rejeject_bef_you_run_results}, where they find good prediction quality (as measured by Brier score) from both $\hat{p}$ and self-estimation in predicting performance of all base models. While self-estimation is slightly better, the external alt predictor $\hat{p}$ does not need to run GPT-3 when the rejection rule is enabled, saving computational cost. By rejecting those instances that were predicted with an estimated probability of success lower than 1\%, 46\% of the failures were avoided, at the cost of only rejecting 1.5\% of correct answers. 
 They also report that various meta-features of the task instances and architectural details of the base systems can augment the predictive power of $\hat{p}$, highlighting the relevance of including features beyond what the original task (base problem) considers \citep[Table 3]{zhou2022reject}.}

\begin{table}[h]
\centering
\begin{tabular}{lccc}
\hline
Base model & Base model's Acc.↑ & BS of self-estimation↓  & BS of $\hat{p}$↓ \\
\hline
GPT-3 Ada 350M & 0.524$\pm$0.232  & 0.122 & 0.144 \\
GPT-3 Babbage 1.3B & 0.580$\pm$0.240  & 0.116  & 0.141\\
GPT-3 Curie 6.7B & 0.625$\pm$0.244 & 0.108  & 0.130 \\
GPT-3 Davinci 175B & 0.689$\pm$0.253  & 0.096 & 0.125 \\
\hline
\end{tabular}
\vspace{3pt}
\caption{\inserted{Comparison between Brier Score (BS) of the assessor's predictive power and the self-estimation baseline from GPT-3 models \citep{zhou2022reject}. The average accuracies (with std. dev) across different numbers of shots from base models, GPT-3 variants, on the data-wrangling task are also presented.}}
\label{tab:rejeject_bef_you_run_results}
\end{table}

\begin{figure}[!ht]
    \centering
    \includegraphics[width=0.42\linewidth]{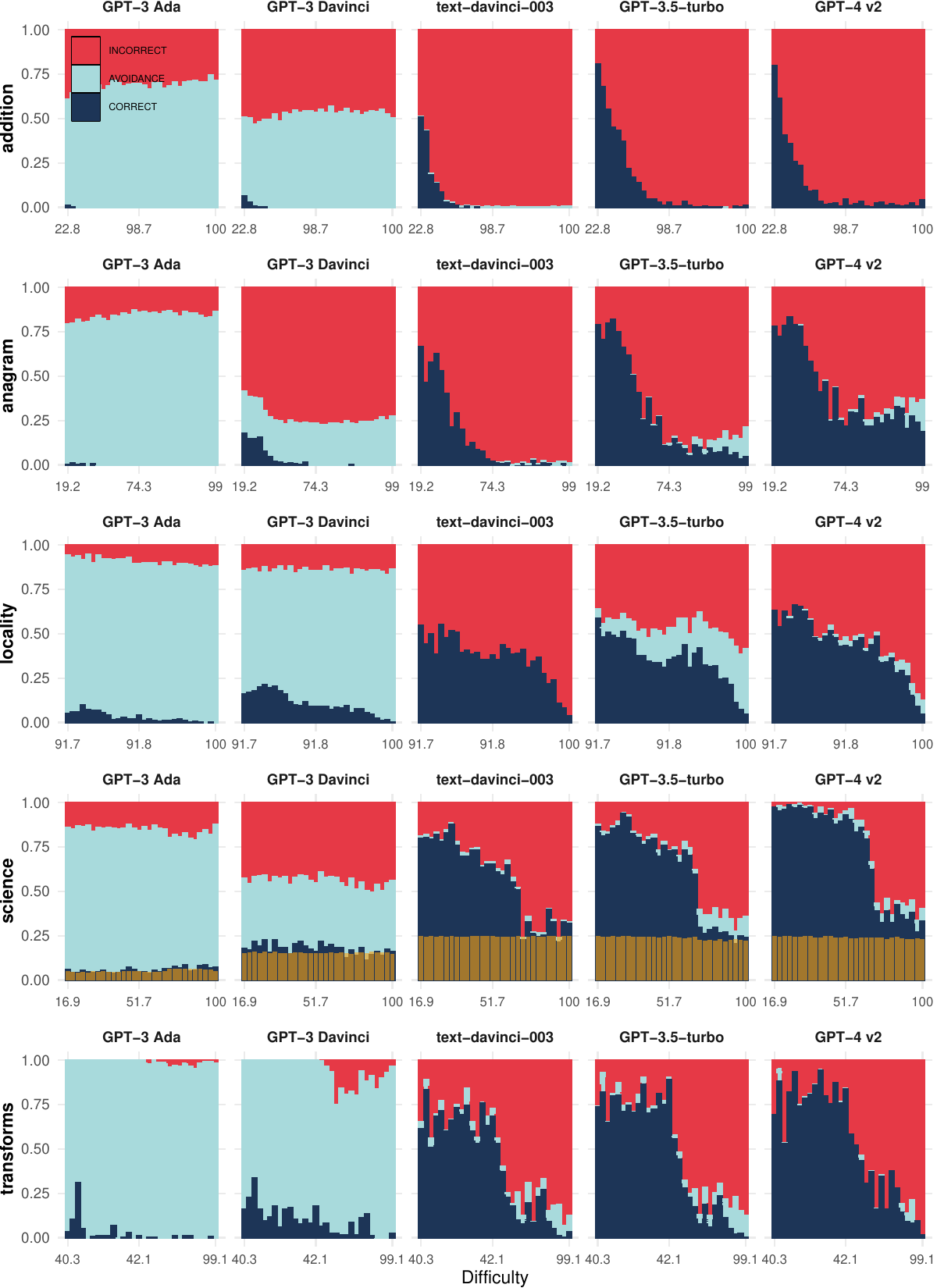}
    \includegraphics[width=0.42\linewidth]{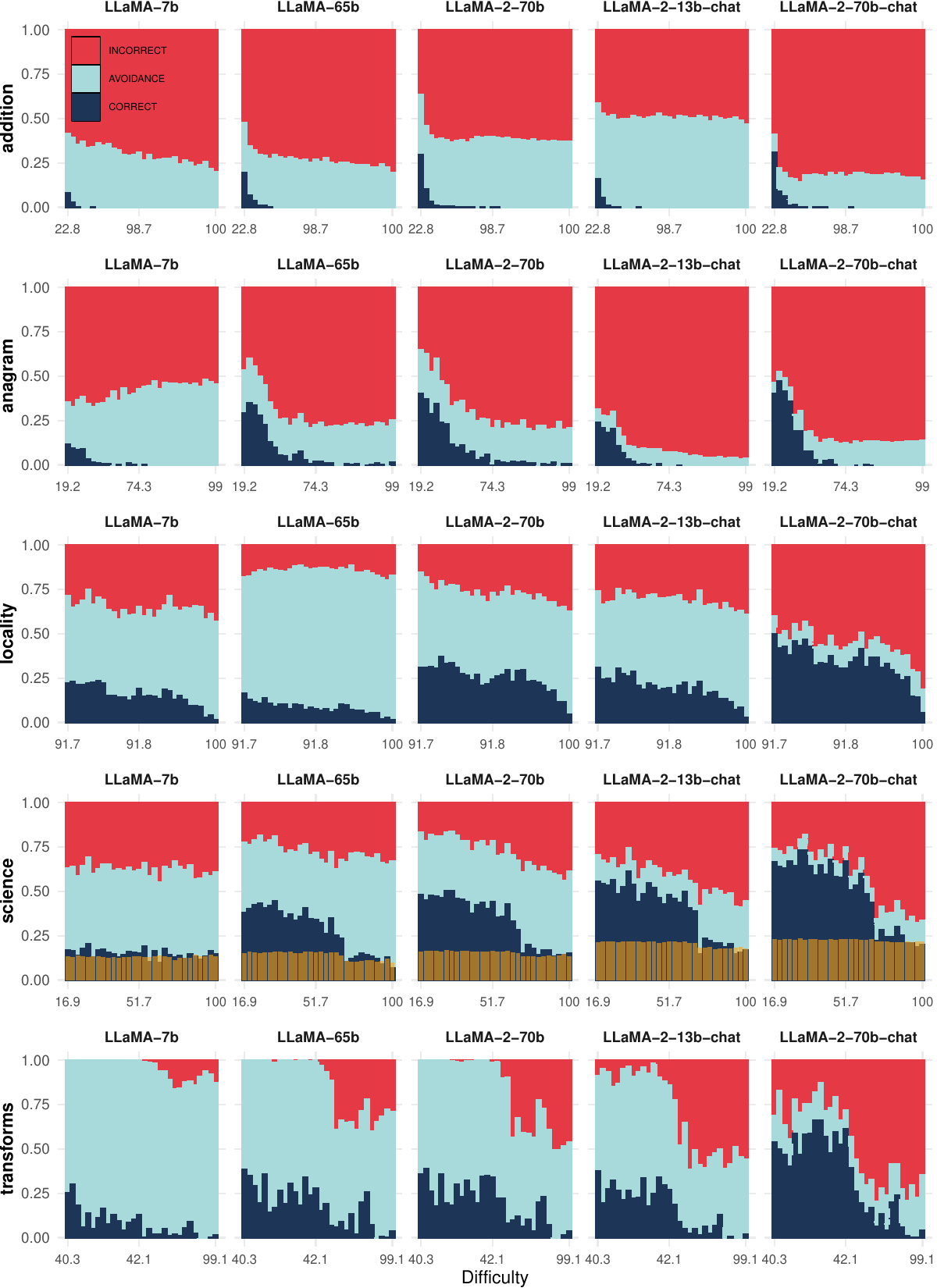}
    \caption{\inserted{Performance of a selection of GPT and LLaMA models over human difficulty on the ReliabilityBench benchmark \citep{zhou2024llmrel}. The values are split by correct, avoidant and incorrect results. The {$x$-axis}\xspace is split into 30 equal-sized bins, whose ranges must be taken as indicative of different distributions of perceived human difficulty across benchmarks.}}
    \label{fig:llmrel_fig2}
\end{figure}

\inserted{In a third scenario, we attempt to explore how easy it is to find features for $\langle i,s,u \rangle$ that are predictive, and how good humans are in finding them. \citet{zhou2024llmrel} found that human-estimated difficulty is a good predictor of performance in LLMs (Figure \ref{fig:llmrel_fig2}). This indicates that future LLMs could use human difficulty to determine when to abstain from providing an answer \citep{brahman2024art}. Furthermore, humans can also use it to reject the model’s output for difficult questions, acting as a predictor. While this is promising for both machine and human oversight, \citet{zhou2024llmrel} noted that in practice humans do not leverage this difficulty well when spotting and rejecting possible errors, corroborating previous observations about humans failing to determine where LLMs fail \citep{carlini_gpt4_challenge}. 
However, the predictability is there, ready to be exploited.}

\begin{figure}[ht]
    \centering
    \includegraphics[scale=0.305]{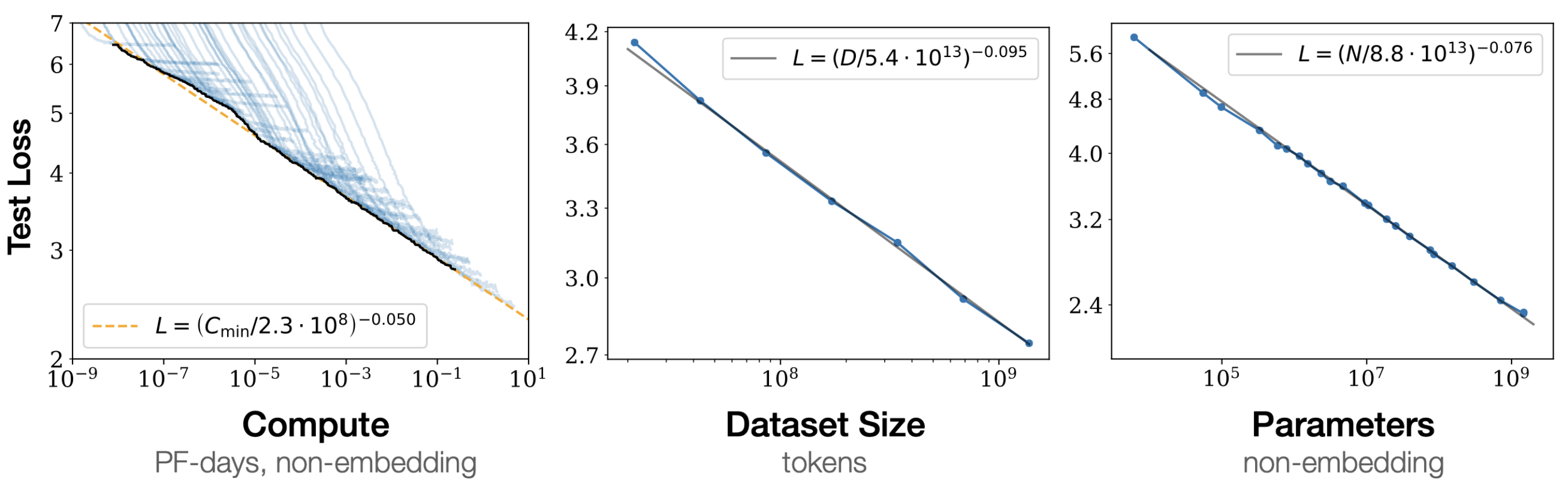}
    \caption{Scaling laws of neural models~\citep{kaplan2020scaling}. The test loss is predictable from the compute used during training, the training dataset size and the number of parameters of the model.}
    \label{fig:scaling_laws}
\end{figure}

Our \modified{fourth} 
scenario focuses on the so-called “scaling laws”~\citep{kaplan2020scaling}, which represent a power-law relationship between the overall performance of language models for a set of tasks and the increase in factors such as model size, dataset size and computational power (see Figure \ref{fig:scaling_laws}). Here, the input variables are compute, data size and number of parameters. These are proven to be highly predictive for neural models’ test loss, with loss linearly decreasing with these parameters (log scale) \citep{kaplan2020scaling, hernandez2021scaling}.

\inserted{This scenario is a clear case of long-horizon hypotheticality that is usually addressed with coarse granularity (how a hypothetical model will perform on a dataset), but there is increasing interest in building predictors at the instance level \citep{schellaert2024proposal}, especially as new models such as OpenAI o1 can be parametrised by the `thinking budget', as they ``improve with more thinking time” \citep{zhong2024evaluation}. 
If we can anticipate that the model can give us a good result by thinking during 5 seconds, why should we give it 100 seconds? Conversely, if we anticipate it cannot give us a good answer, why should we spend all these costly thinking seconds?}

These scenarios emphasise the relevance of the input features and also share an anticipative character. They differ on the prediction horizon, and are situated at the two extremes of aggregation: the local, fine-grained prediction at the instance level \inserted{(for the first three scenarios)}; versus the global, coarse prediction for massive benchmarks \inserted{(the fourth scenario)}. These extremes suggest there are many intermediate areas where predictability has not been explored. These 
\inserted{four} examples also highlight the difference between predicting performance of a specific AI system and making a more general prediction about a class of hypothetical \modified{(not yet trained)  AI systems.} This exploration of intermediate levels, varying scales and 
\modified{different types of validity indicators} is fundamental to understanding possibly confounding effects of the aggregation, such as a biased selection of the relevant input or output variables until predictability is found~\citep{schaeffer2024emergent}.

\section{The trade-offs}\label{ssec:tradeoffs}

\inserted{
We advocate for a paradigm shift where the design, selection and use of predictable AI systems is prioritised. However, there is a tension between predictability and the quality of the base systems, because a model that always fails is fully predictable. There is further tension in the effort that must be expended to minimise Eq.~\ref{eq:ecosystemunpred}. Ultimately,  what we would like is a set of 
useful systems $s_A, s_B, ... \in {\sf S}$ and a good predictor $\hat{p}$ for the resulting validity indicators of the AI ecosystem. 
}

\inserted{
The first idea for building more predictable AI systems, especially machine learning models, may be based on keeping them simple (e.g., a set of causal rules instead of a complex black box model). However, this only entails behavioural predictability but may not ensure more validity predictability. For instance, in a classification problem, an AI system $s$ that always predicts the same label is very simple and very explainable, but predicting where it fails, the $\hat{p}$ problem, would require learning the original classification problem. In general, if the AI models ${\sf S}$ have not captured the epistemic uncertainty of the base problem, this will make the task of finding $\hat{p}$ harder, as this epistemic uncertainty would need to be instead captured by $\hat{p}$.
}

\inserted{Following the balance between minimising $\unpred$ and maximising expected validity $\mathbb{V}$ we initially explored in Eq.~\ref{eq:envelope}, here we point out some other ways to interpret this trade-off:}

\begin{itemize}
\item \inserted{Explore the Pareto between the expected validity $\mathbb{V}$ and reducing the loss $S$ of $\hat{p}$ w.r.t. $p$. 
For instance, in the scenario of the Animal AI Olympics seen above, there were some participants, such as `Sparklemotion’,  
that showed higher accuracy than other weaker participants, such as `Juohmaru’, but much worse predictability \citep{burnell2022not}. a A Pareto plot ($x$-axis and $y$-axis equal to accuracy and predictability, respectively) could place `Juohmaru' as preferable. 
}

\item \inserted{In the case of binary validity distribution $p$, if the predictor $\hat{p}$ is well calibrated, we could set a threshold 
to determine the proportion of operating conditions ${\cal E}_{\hat{p},\tau} = \{ e \in {\cal E} : \hat{p}(V=1\:|\:e) \geq \tau\}$ that are not rejected and the \% of these regions that are actually above the threshold of quality, i.e., $\mathbb{E}_{e \in {\cal E}_{\hat{p},\tau}} p(V=1\:|\:e) \ge \tau$. In the second scenario, \textit{“E7. cost-effective data wrangling automation”}, the proportion of operating conditions increased from 55.2\% (without rejection) to 69.2\% at the cost of rejecting 1.5\% of correct answers, with $\tau = 0.01$\footnote{\inserted{This is illustrated in Table 4 of \citep{zhou2022reject}.}}. If the predictor is well calibrated, this proportion can be further increased by increasing $\tau$, but this will also further reduce the number of correct answers the user receives.} 

\item \inserted{Instead of setting a threshold, which is very application or context-dependent, we can optimise for some metrics that combine high validity and low rejection using $\hat{p}$, such as the area under the accuracy-rejection curve \citep{condessa2017performance} or extensions beyond classification.} 
\end{itemize}


\inserted{Specific solutions can be tailored for each AI ecosystem, but the choice of the validity metric, its maximisation (through better AI systems) and the optimisation of its predictability (through better predictors $\hat{p}$) will be central to the essential challenges and opportunities of Predictable AI.}

 \section{Challenges and opportunities}

 Characterising the \modified{field} of Predictable AI allows us to better delineate its challenges and turn them into focal research opportunities rather than scattered efforts. The following list is not exhaustive, but builds on the elements identified in previous sections:

 \begin{itemize}
     \item \textbf{Metrics}: Can we use the traditional evaluation metrics for performance, usefulness, safety, etc., or do we need new metrics such as alignment, honesty, harmlessness, helpfulness~\citep{askell2021general}? \modified{What properties make a metric more easily predictable? How do we identify when a system is predictable enough?}
     
     \item \textbf{Evaluation data}: What data to collect for training predictive models or evaluate their predictiveness?~\citep{burnell2023rethink} How can we combine human feedback, predictions from different actors~\citep{carlini_gpt4_challenge}, results from other systems~\citep{liang2022holistic}, incident databases~\citep{toner2021incident}, meta-feature construction and annotations~\citep{gilardi2023chatgpt}?
     
     \item \textbf{Aggregation and disaggregation}: Can different predictability problems at several granularities be bridged, from local, instance-level predictions to global, benchmark-level predictions and vice versa? Is quantification~\citep{esuli2023learning} the right tool for this?

          \item \moved{\textbf{Effective monitoring}: How can we integrate different predictors to monitor AI ecosystems and federate them~\citep{li2020federated} in case of multiple users and stakeholders? What are the liability implications and how should this be regulated?}
          
     \item \textbf{Reuse of knowledge}: How can we reuse domain knowledge from cognitive science about how humans and animals solve tasks~\citep{lake2017building,momennejad2023rubric,crosby2019translating,crosby2020animal} or from what explainable and interpretable AI finds about an AI system?.

 \end{itemize}

 \inserted{Regarding all these challenges, and especially the reuse of knowledge, we see opportunities in exploiting the synergies of comparing validity predictability with predictability in other sciences  
~\citep{grunberg1954predictability,stern2009genetic,song2010limits,conway2010evolution,kello2010scaling,kosinski2013private,svegliato2018meta,mellers2014psychological,salganik2020measuring,wintle2023predicting}.}

\inserted{A crucial research niche that is intertwined with the previous list is the \textit{identification of pathways toward improving the predictability} of AI ecosystems
. There are several promising methods for this. \cite{zhou2024llmrel} propose two ways to increase error predictability of LLMs from a human perspective: (1) modify loss functions to penalise errors on easy tasks more heavily than difficult ones so as to enhance the concordance between user difficulty expectations and model errors; 
(2) use human expectation to make LLMs more epistemically human-like such as abstaining from answering on tasks beyond their capabilities. In a different approach, \citep{premakumar2024unexpected} demonstrate that neural networks self-regularise when given the auxiliary task of predicting their own internal states, making networks more parameter-efficient and reducing complexity, which may increase their predictability. Further, recent research on mechanistic interpretability using Sparse Auto-Encoders (SAEs) has shown promise in learning monosemantic representations from transformer network layers \citep{bricken2023monosemanticity}. By integrating trained SAEs into model architectures, researchers can intervene on models to change their internal representations, which could be used to optimise for predictability with respect to a given pair of base model and predictor.}

\modified{
Depending on the domain, there are open methodological questions such as who should make the predictions (human experts, the AI systems themselves or an external predictive model) and how their predictions should be elicited. There are further theoretical questions about \textit{how much} can be predicted subject to aleatoric and epistemic uncertainty, and the causal loops involving predictions. There are further ethical issues, such as privacy of behaviour and responsibility when predictions fail. In general, many of the above challenges will lead to cross-disciplinary research opportunities.}

\section{Impact and vision}

 We identified AI predictability as a fundamental, \modified{yet} underexplored component of \inserted{many} AI desiderata, \inserted{situating it as a field of scientific enquiry in itself}. \modified{Predictability is highly intertwined with, but separate from, other important factors such as safety, alignment, explainability, trust, liability and control.}  \inserted{A collective shift in focus towards Predictable AI would constitute a profound paradigm shift yielding greater assurances about system performance, safety and deployment suitability.} 
 \modified{There are reasons to be optimistic about predictability within AI: first, in many other sciences, predictability is a fundamental aspect of operation, and many ideas can be reused;} \inserted{second, there has been enormous progress in predictive techniques, and we expect foundation models to be used as predictors of validities in many domains, as we have seen with LLMs.} 
 
 \modified{Until now, the field of Predictable AI had not been properly defined and is still vastly under-explored. \inserted{We anticipate that through the framework presented in this paper, concrete progress can now be made.} In particular, the use of machine learning to exploit the increasingly large amounts of evaluation data (benchmark results and human feedback) generated by AI systems holds promise for the development of this nascent field, leading to a landscape of predictably valid AI systems.}


\bibliographystyle{plainnat}
\bibliography{refs.bib}
\end{document}